\documentclass[letterpaper, 10 pt, conference]{ieeeconf}  % Comment this line out if you need a4paper

\IEEEoverridecommandlockouts                              % This command is only needed if 
\usepackage{graphics}
\usepackage{epsfig}
\usepackage{mathptmx}
\usepackage{times}
\usepackage{amsmath} 
\usepackage{amssymb}
\usepackage{booktabs}
\usepackage{cite} 
\usepackage{multicol}
\usepackage{float}
\usepackage{multirow}
\usepackage{ulem}
\usepackage{hyperref}
\usepackage[hypcap=false]{caption}
\newcommand{\poseit}{\textit{PoseIt}}
\newcommand{\datapoints}{1840}

\title{\LARGE \bf\poseit: A Visual-Tactile Dataset of Holding Poses\\ for Grasp Stability Analysis}
\author{ Shubham Kanitkar$^{*, 1}$  \hspace{1cm} Helen Jiang$^{*, 1}$  \hspace{1cm} Wenzhen Yuan$^{1}$\\
\thanks{$^{*}$Authors with equal contribution.}
\thanks{$^{1}$Carnegie Mellon University}
\thanks{\texttt{\{skanitka, helenjia, wenzheny\}@andrew.cmu.edu}}}
\begin{document}

\thispagestyle{empty}
\pagestyle{empty}

\makeatletter
\let\@oldmaketitle\@maketitle%
\renewcommand{\@maketitle}{\@oldmaketitle%
    \centering
    \includegraphics[width=\linewidth]{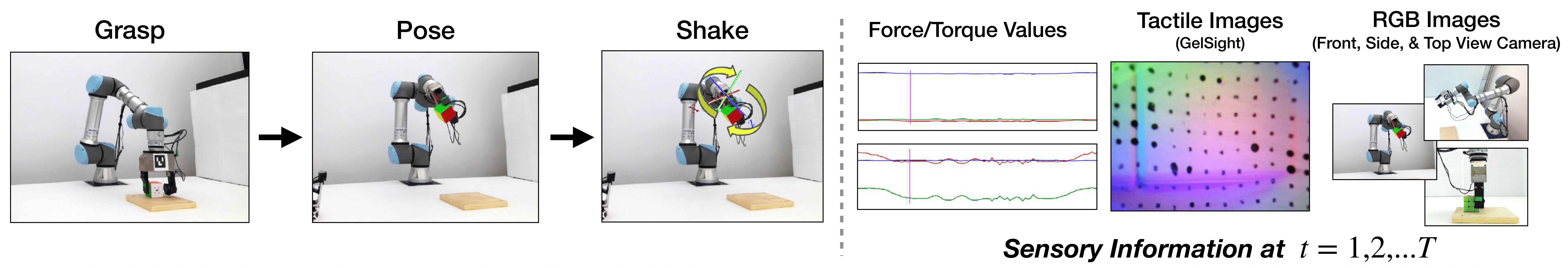}
    \vspace{-0.1in}
     \captionof{figure}{The~\poseit~data collection pipeline contains the following phases: grasping the object, moving the object to a holding pose, and shaking the object to check its stability. During the process, we collect RGB-D, tactile, force, torque, joint angle, joint velocity, and gripper force information. In total, \poseit~consists of \datapoints~grasp datapoints collected from 26 distinct objects, with 16 different holding poses for each object.}
     \label{teaser}
}
\makeatother
\maketitle
%%%%%%%%%%%%%%%%%%%%%%%%%%%%%%%%%%%%%%%%%%%%%%%%%%%%%%%%%%%%%%%%%%%%%%%%%%%%%%%%

%%%%%%%%%%%%%%%%%%%%%%%%%%%%%%%%%%%%%%%%%%%%%%%%%%%%%%%%%%%%%%%%%%%%%%%%%%%%%%%%
\begin{abstract}
When humans grasp objects in the real world, we often move our arms to hold the object in a different pose where we can use it. In contrast, typical lab settings only study the stability of the grasp immediately after lifting, without any subsequent re-positioning of the arm. However, the grasp stability could vary widely based on the object's holding pose, as the gravitational torque and gripper contact forces could change completely. To facilitate the study of how holding poses affect grasp stability, we present~\poseit, a novel multi-modal dataset that contains visual and tactile data collected from a full cycle of grasping an object, re-positioning the arm to one of the sampled poses, and shaking the object. Using data from~\poseit, we can formulate and tackle the task of predicting whether a grasped object is stable in a particular held pose. We train an LSTM classifier that achieves 85\% accuracy on the proposed task. Our experimental results show that multi-modal models trained on~\poseit~achieve higher accuracy than using solely vision or tactile data and that our classifiers can also generalize to unseen objects and poses. 

The~\poseit~dataset is publicly released here: \url{https://github.com/CMURoboTouch/PoseIt}.
\end{abstract}
%%%%%%%%%%%%%%%%%%%%%%%%%%%%%%%%%%%%%%%%%%%%%%%%%%%%%%%%%%%%%%%%%%%%%%%%%%%%%%%%

%%%%%%%%%%%%%%%%%%%%%%%%%%%%%%%%%%%%%%%%%%%%%%%%%%%%%%%%%%%%%%%%%%%%%%%%%%%%%%%%
\section{Introduction}
Grasping is a core component of many complicated manipulation tasks in robotics. Traditionally, research in grasping focuses on detecting grasping locations~\cite{miller2003automatic,saxena2008robotic,redmon2015real}, and maintaining the grasp stability~\cite{romano2011human,bekiroglu2011assessing,bekiroglu2011learning,su2015force,calandra2018more}. These prior works focus on a setting where the gripper holds the object vertically in the robot's hands. Stability is only evaluated in a fixed pose after the robot lifts the object. 
 
This does not translate well in the real world where humans rarely hold the object perfectly still immediately after lifting --- for functional purposes, we often need to move the object to a different pose.
 However, the stability of the grasp can vary significantly with the pose, which is a shortcoming of prior settings which only study the pose immediately after lifting. For example, if a sword is held with its blade pointing vertically to the sky, gravity doesn't create any torque on the sword. If the blade runs parallel to the ground, the torque from gravity could cause it to slip, which is potentially dangerous. 

We use ``\textit{holding pose}" to describe the pose of the object when it is held in the gripper. Humans have the ability to select a holding pose that is both stable and appropriate for using the object. Humans use the ``feeling'’ from fingers to quickly understand whether the current pose is a good one or if the object is at risk of slipping. The key insight is that the tactile information enables this capability as opposed to using solely the visual signals. We believe robots could work in a similar way, by using both tactile and visual feedback to evaluate holding poses for objects.

In this paper, we propose a data-based method for predicting the grasp stability of objects in different holding poses. We build a visual-tactile dataset,~\poseit, to record the sensory feedback when a robot with a parallel-jaw gripper grasps the object, moves to a holding pose, and shakes the object. We label whether the grasp was stable during each of these steps and collect data for 26 different objects at 16 different poses. We aim to create a comprehensive dataset with many different modalities of sensory information that the community can refer to when studying holding poses. To collect tactile data, we use a high-resolution GelSight sensor~\cite{yuan2017gelsight} on the fingers. To collect the visual data, we set three cameras to record the grasping point, object geometry, and the overall view of the robot’s and object’s motion.  We also use a force-torque sensor on the robot’s wrist.

\addtocounter{figure}{-1}

\begin{figure}[ht]
\centering
\includegraphics[width=\linewidth]{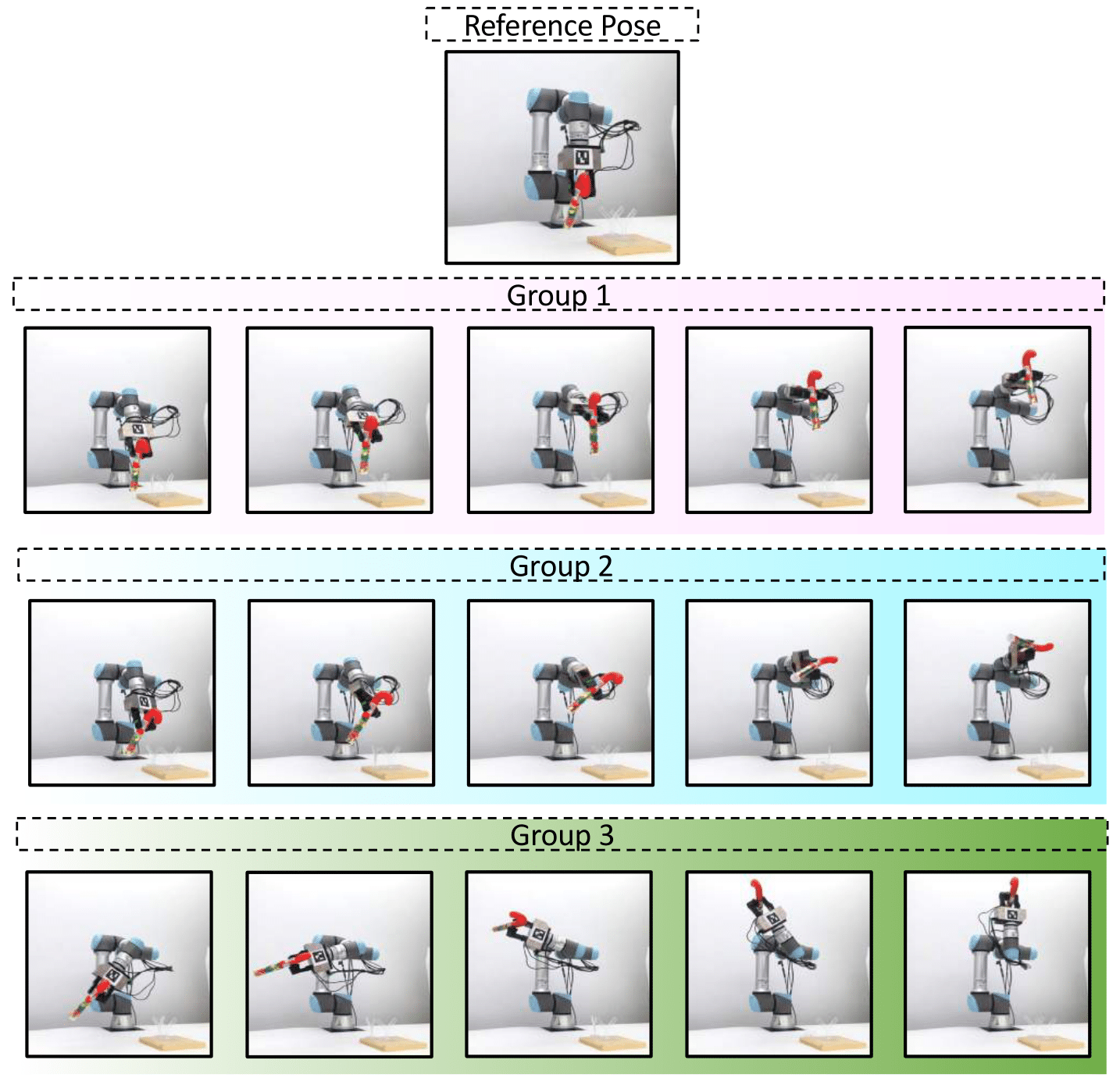}
\caption{\textbf{Holding pose sample space.} An example of the 16 sampled poses. Each trial starts at the reference pose, which is the base position where the gripper faces downwards. We create 3 groups of 5 holding poses each, based on the gripper’s final orientation.}
\label{fig:grasp_pose_sample_space}
\end{figure}

\begin{figure}[ht]
    \centering
    \includegraphics[width=\linewidth]{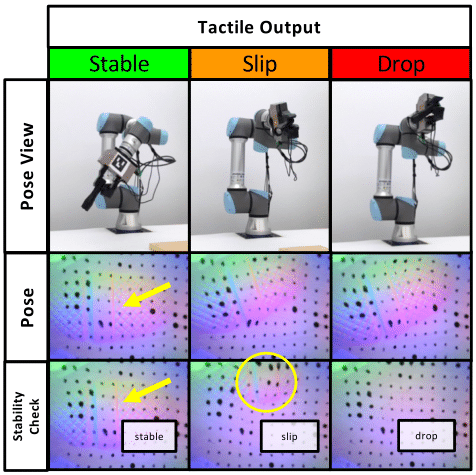}
    \caption{The tactile data for a flashlight grasped at the tail using 80N force. For the stable pose in the left column, the grasp is stable where the flashlight is clearly imprinted during the stability check. In the middle column, the flashlight slips during the shaking phase -- this is visible where the imprint of the object attenuates gradually and moves toward the top of the sensor surface. Whereas in the pose where the flashlight dropped in the right column, the imprint on the sensor surface disappears during stability check.}
    \label{fig:flashlight}
\end{figure}

We use data from~\poseit~to formulate and tackle the task of predicting grasp stability in a particular holding pose. We use tactile and vision data obtained during grasping and moving the object to the holding pose to predict whether the object is stable. In contrast, most prior work focuses on studying stability in a single pose immediately after the object is lifted.  

An important and challenging requirement for solving our task is to correctly classify cases where the object appears stable but will slip if the robot shakes it. Such cases are in the minority ($\approx 20$\% of the full dataset), but important to accurately detect in practical scenarios. To this end, we train an LSTM classifier on visual and tactile data from~\poseit~using techniques for learning with imbalanced datasets~\cite{cao2019learning}. 

Our classifier trained on more poses using tactile and vision data achieves 85.2\% accuracy on held-out unseen poses, which is 3.4\% better than a model which trains on fewer unique poses (Section~\ref{sec:unseen_poses}). This demonstrates that a diverse set of holding poses for each object can improve generalization to unseen poses. We also found that using tactile and vision together is 13.2\% better than using vision alone and 3.4\% better than using tactile alone, demonstrating the value of collecting multi-modal data. 

In summary, we have two primary contributions: 1) we propose~\poseit, a novel dataset with multi-modal tactile and visual information and labels for the stability of an object through various stages of grasping, moving to a holding pose, and shaking. 2) we use~\poseit~to train a classifier which predicts whether the grasp is stable in the current holding pose.
%%%%%%%%%%%%%%%%%%%%%%%%%%%%%%%%%%%%%%%%%%%%%%%%%%%%%%%%%%%%%%%%%%%%%%%%%%%%%%%%

%%%%%%%%%%%%%%%%%%%%%%%%%%%%%%%%%%%%%%%%%%%%%%%%%%%%%%%%%%%%%%%%%%%%%%%%%%%%%%%%
\section{Related Work}
\begin{figure*}[ht]
    \centering
    \includegraphics[width=\linewidth]{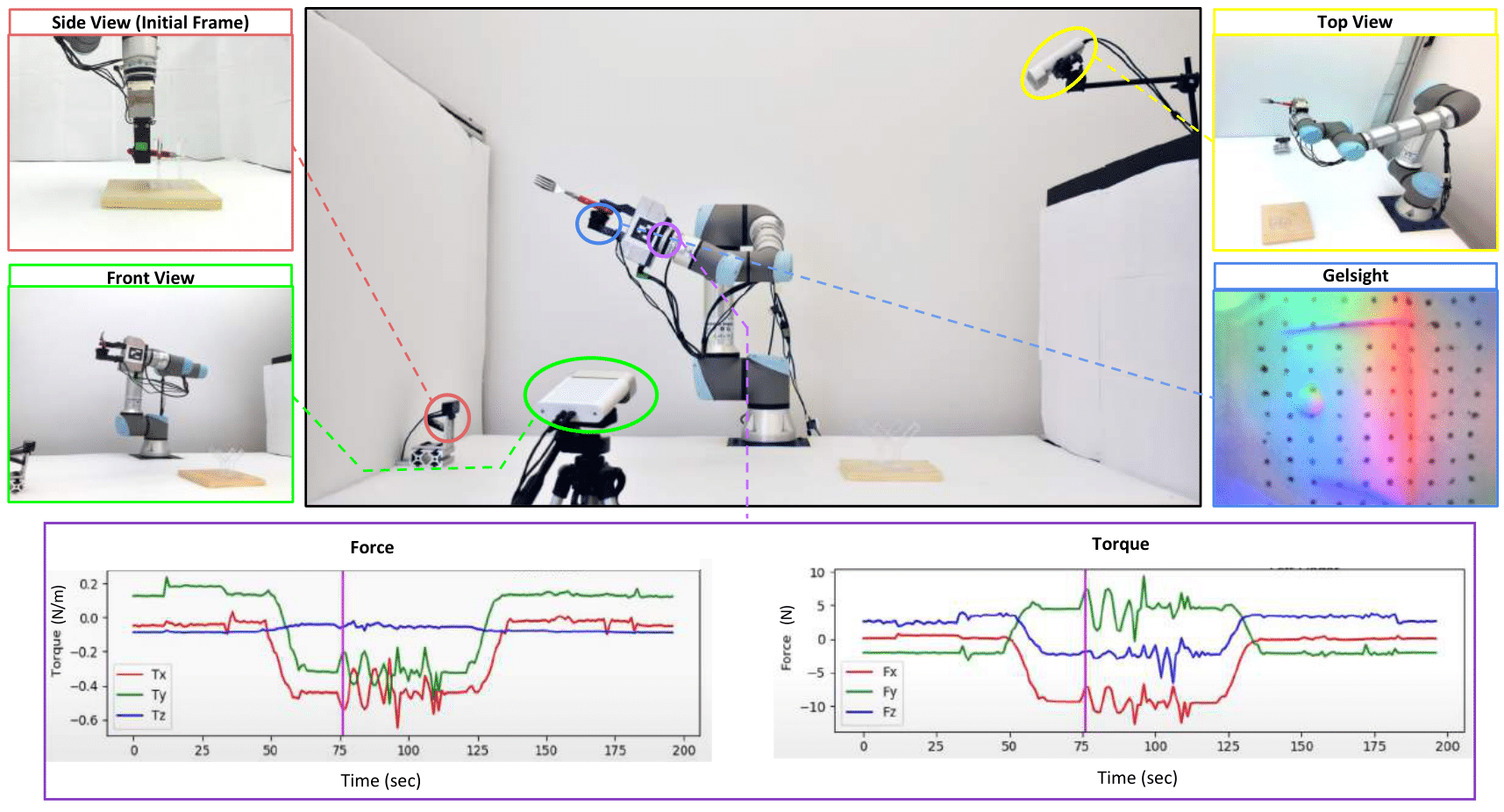}
    \caption{ {\bf Robot Setup and Data Modalities.} We use this setup to collect~\poseit, which consists of multi-modal data on 26 objects, including RGB-D cameras, GelSight tactile sensor, force/torque sensor, and robot trajectory.}
    \label{fig:data_modalities}
\end{figure*}

The grasping literature can be roughly categorized into two styles of approach: analytic and data-driven~\cite{sahbani2012overview,bohg2013data}. Analytic approaches rely on known physical models of the object, environments, and grippers to construct the grasps and reason about their quality~\cite{nguyen1987constructing,shimoga1996robot,li2003computing,sahbani2012overview}. However, these approaches could fail if the correct modeling assumptions are unknown or misspecified. Our work is more related to the long line of data-driven approaches, which rely on observations of past trials to build a model or classifier for grasping. Prior works in the data-driven category vary widely in terms of the data modalities and modeling techniques that they use~\cite{goldfeder2009data,bekiroglu2011learning,bekiroglu2013probabilistic,bohg2013data,levine2018learning,pinto2016supersizing,guo2016deep,li2018slip}.

\noindent{\bf Predicting grasp stability.} Our work is most closely related to papers which use tactile data for data-driven grasp stability prediction~\cite{bekiroglu2011assessing,madry2014st,calandra2017feeling,li2018slip,calandra2018more,veiga2018grip,zhang2018fingervision,zapata2019learning}. A common theme in these papers is to use tactile data collected from multiple trials of grasping and lifting to predict stability after lifting (and possibly adjust the grasp).~\cite{bekiroglu2011assessing}~use machine learning models such as Support Vector Machines (SVMs) and Hidden Markov Models (HMMs) to predict slip based on data from a Weiss robotics tactile sensor.~\cite{calandra2017feeling} predicts slip using a neural network trained on visual and tactile data from the GelSight sensor, and in follow-up work~\cite{calandra2018more}, combine slip prediction with an action-conditional model to learn grasping and re-grasping sequences in an end-to-end manner.~\cite{li2018slip} and~\cite{zhang2018fingervision} also train neural networks using visual and GelSight data for the task of slip detection as the object is lifted. All of these prior works consider the stability of the object in the pose obtained immediately after grasping. In contrast, the primary contribution of our work is to collect visual and tactile data for grasping and moving the object to a diverse set of holding poses, as objects can behave differently in these holding poses than the pose obtained after lifting.  This allows us to tackle the (to the best of our knowledge, novel) task of predicting grasp stability in a particular holding pose.

\noindent{\bf Grasping datasets.} Large-scale data is a crucial driving force behind many of the recent advancements in grasping. For example,~\cite{pinto2016supersizing} collect a dataset of 50000 grasp trials, which was 40x larger than prior work, and show that this dataset can be used to predict grasp locations from image patches (they do not use tactile sensing).~\cite{chebotar2016bigs} collect a dataset of 1000 grasp trials using BioTac sensors~\cite{wettels2008biomimetic} and perform a shaking-based stability check to determine whether a grasp is stable. The datasets of~\cite{calandra2017feeling,li2018slip}, which contain GelSight and visual data, are also publicly available, with 9269 and 1102 grasp trials, respectively.~\cite{murali2018learning} collect a dataset of 7800 grasp interactions involving localizing the object, grasping, and regrasping and train an iterative regrasping policy based on tactile feedback.~\cite{Wang2019MultimodalGD} collect a dataset of 2550 grasp trials with the Eagle Shoal robotic hand. While there are now many choices of publicly available tactile datasets for initial grasp stability, prior to our work, no dataset existed for evaluating grasp stability in various holding poses.
%%%%%%%%%%%%%%%%%%%%%%%%%%%%%%%%%%%%%%%%%%%%%%%%%%%%%%%%%%%%%%%%%%%%%%%%%%%%%%%%

%%%%%%%%%%%%%%%%%%%%%%%%%%%%%%%%%%%%%%%%%%%%%%%%%%%%%%%%%%%%%%%%%%%%%%%%%%%%%%%%
\section{Collecting the poseit dataset}
In this section, we describe the collection process for the~\poseit~dataset. Two main components of~\poseit~differ from prior works: 1) the holding pose sample space, a set of 16 distinct holding poses to mimic typical human-like holding poses 2) the multi-phase collection cycle, which moves an object to a particular pose and performs a shaking stability check at that pose. 

\subsection{Holding pose sample space}
The stability of a particular object in the gripper depends on the holding pose. To attempt to model how humans typically hold objects for use (rather than simply lifting them up vertically), we design a diverse set of 16 holding poses.

We set the first pose as a reference pose, as it is the typical base position where the gripper faces downwards. Inspired by human arm movements observed during manipulating routine objects, the other 15 holding poses are categorized into 3 possible groups of 5 poses, each based on the final orientation of the gripper.
We expect poses within the same group to behave more similarly.  We obtain the poses within a group by rotating the end-effector in fixed-degree increments. Examples of the 16 holding poses are shown in Figure~\ref{fig:grasp_pose_sample_space}.
Group 1 poses take advantage of gravity to stabilize potential rotational movements in the case of objects with varying mass distributions. Whereas, poses in group 2 gradually rotate the gripper to counterbalance the gravitational force to avoid slippage for objects with non-uniformly distributed mass. Group 3 poses aid in analyzing the stability of the overhead (above the shoulder joint) arm movements by incrementally increasing the object's distance from the ground. 

\begin{figure}[ht]
    \centering
    \includegraphics[width=0.9\linewidth]{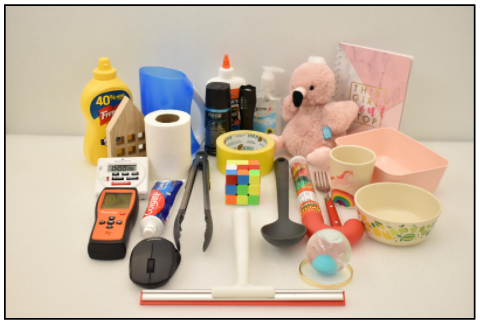}
    \caption{ {\bf Dataset Objects.} To test grasp quality on a diverse range of objects, we collected data for 26 various household objects with a diverse range of size, shape, material, mass, and texture. For each object, we grasp it at multiple locations while it is resting in most likely orientations. We also grasp each object with 2 different gripper forces, the minimum force required to lift the object from (5N, 15N, or 40N), and 80N. We collect data for all possible combinations of grasp point, force value, and target holding pose, resulting in (Number of grasp points~$\times2\times16$) choices per object.}
    \label{fig:objects}
\end{figure}

\begin{figure*}[ht]
    \centering
    \includegraphics[width=16cm]{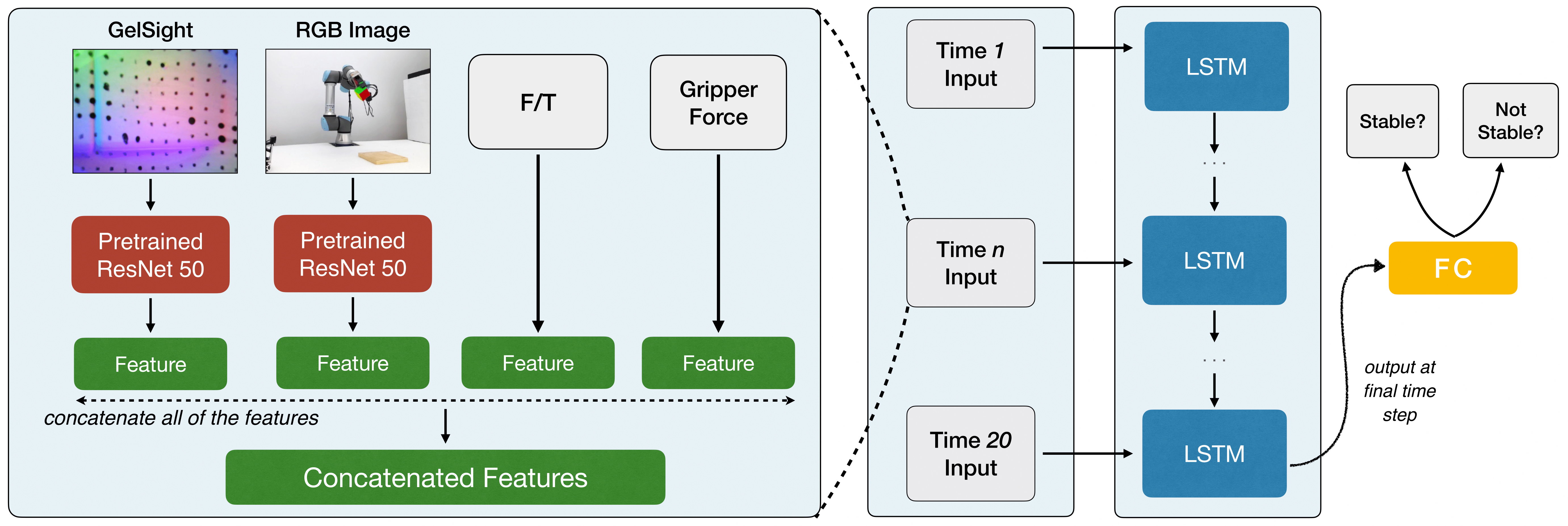}
    \caption{\noindent{\bf Model architecture for grasp stability prediction.} Our model takes as input 4 modalities of data: tactile, visual, F/T sensor data, and the gripper force value (which is fixed across all timesteps). We use the sequence of concatenated features from these modalities as input to an LSTM, which predicts ``stable'' or ``not stable''.}
    \label{fig:model}
\end{figure*}

\subsection{Data collection pipeline} 
At the start of every data collection cycle, the object is placed at the center of the table in varying orientations. It is then grasped and moved through 4 phases in succession:
\begin{enumerate}
\item Grasp: The robot grasps and lifts the object. Data collection begins 1 second before grasping.
\item Pose: After lifting the object, the robot arm moves to one of 16 possible holding poses. The robot waits for 1 second after reaching the pose. 
\item Shaking/stability check:
 This phase shakes the object in-hand, inspired by prior work who also performs in-hand shaking~\cite{chebotar2016bigs,wang2020swingbot}. We apply a quick rotational movement along the gripper axis, followed by rigorous arm shaking movements along all 3 axes. This phase tests whether the object is stable in the holding pose. 
\item Retract and release: During the retract stage, the robot arm returns to the beginning position, and the gripper releases the object at its starting location. 
\end{enumerate}

\noindent{\bf Data annotation.} We manually label the stability of each of the 4 phases with one of 4 categories:
\begin{itemize}
    \item Pass: Firm grasp. The object doesn't move relative to the gripper.
    \item Slip: Object is in still contact with the gripper, but some rotational or translational slip is manually observed. 
    \item Drop: Object falls off the gripper.
    \item Not present: Object dropped before the current phase.
\end{itemize}
\section{Analysis of dataset statistics}
In total,~\poseit~consists of \datapoints~datapoints from 26 diverse objects across 16 different poses.

We show the phase-wise data division in Table \ref{tab:gelsight_data_division}. During the grasp phase, we manually collect the roughly same number of stable (\textit{Pass}) and unstable (\textit{Slip+Drop}) initial grasp cases to form a balanced dataset. We observe that in the pose phase, the number of \textit{Slip+Drop} category samples decreases by 14\% (compared to the grasp phase). This shows that post-grasp arm re-positioning can help stabilize the object. 
In the shaking phase, there is a 4\% increase in slip and drop cases (compared to the pose phase). This conveys that the shaking phase is important because even seemingly stable poses may not withstand external disturbances. 

\begin{table}[h]
\centering

\begin{tabular}{@{}lcccc@{}}
\toprule
\multicolumn{1}{c}{\multirow{2}{*}{Stage}} & \multicolumn{3}{c}{Label}                                                        & \multicolumn{1}{c}{\multirow{2}{*}{Slip+Drop}} \\ \cmidrule(lr){2-4}
\multicolumn{1}{c}{}                       & \multicolumn{1}{c}{Pass} & \multicolumn{1}{c}{Slip} & \multicolumn{1}{c}{Drop} & \multicolumn{1}{c}{}                             \\ \midrule
Grasp                                        & 1037                      & 778                       & 25                        &  \textbf{43.64\%}                                           \\ \midrule
Pose                                         & 1278                      & 192                       & 345                       & \textbf{29.18\%}                                           \\ \midrule
Shake                                    & 1003                      & 255                       & 365                       & \textbf{33.69\%}                                         \\ \bottomrule
\end{tabular}
\caption{\textbf{Dataset statistics.} We collected 1840 data points on 26 objects with stability labels on different stages. The last column shows the percentage of unstable cases.}
\label{tab:gelsight_data_division}
\end{table}
\begin{figure}[ht]
    \centering
    \includegraphics[width=0.48\textwidth]{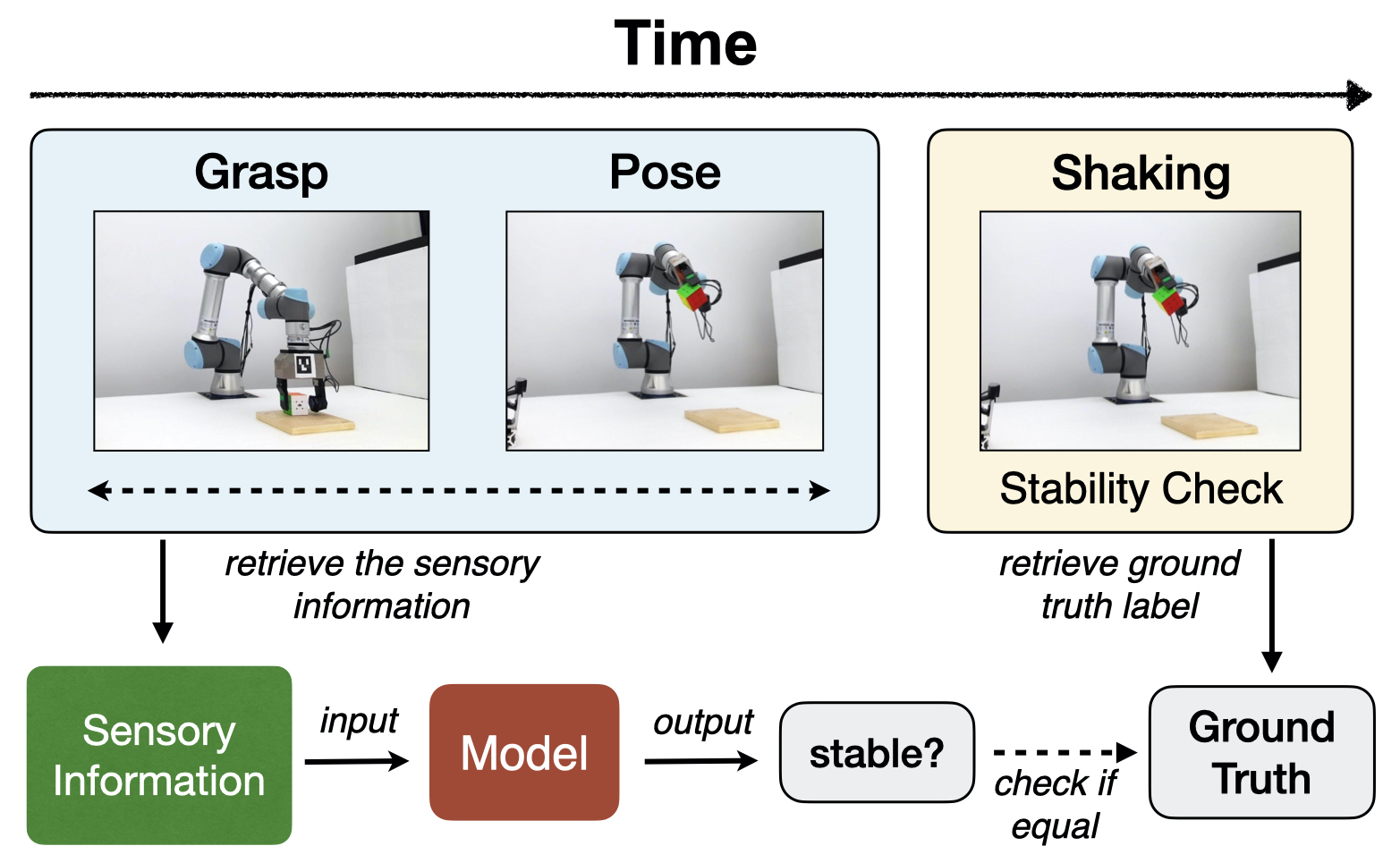}
    \caption{\noindent{\bf Task definition.} Our task aims to predict the stability of the grasp in the holding pose. Given sensory information in the grasp and pose stage, we aim to predict whether the object will be stable during the shaking phase.}
    \label{fig:task}
\end{figure}

\section{Predicting stability in the holding pose} \label{sec:task}
Using the data collected from~\poseit, we predict the grasp stability when the object is in the holding pose. More formally, we formulate our task as follows:
\begin{center}
    \it Given sensory data from the grasp and pose phase, predict whether the object is stable during the shaking phase.
\end{center}

Our task is illustrated in Figure~\ref{fig:task}. The motivation is that understanding and predicting the stability of objects in holding poses is important for manipulating objects in practical settings. We note that our task is a prediction, and not a detection task because the inputs are sensory information before the shaking occurs. 

For simplicity, we combine the ``slip'' and ``drop'' labels so the task is to predict ``stable'' or ``not stable''. From now on, ``label'' denotes the binary label unless specified otherwise. 

\subsection{Model architecture} \label{model_arch}

As in~\cite{li2018slip}, our primary classification model is a Long Short-Term Memory (LSTM) which aggregates multi-modal sensor data over a sequence of time steps to obtain a prediction. Figure~\ref{fig:model} illustrates the model. For a single example, the input is tactile, RGB, and force/torque sensor readings gathered during a sequence of timesteps. The gripping force is also an input, but it is fixed across all timesteps.

To obtain features for the tactile data, we subtract the pre-contact tactile image from all frames in the sequence and feed the resulting frames through a pre-trained ResNet50~\cite{he2016deep} ImageNet backbone. We obtain features for the RGB data using the same pre-trained classifier. We hold the pre-trained classifier throughout training. For a given timestep, we concatenate the tactile features, RGB features, and raw force/torque values for this timestep into a single feature vector. We additionally concatenate the gripper force (a fixed value through all the timesteps) to this vector. The sequence of feature vectors for all timesteps is fed into a 2-layer bidirectional LSTM. The prediction is computed as a linear function of the LSTM hidden layers corresponding to the last timestep in the sequence. 

{\noindent \bf Data features and pre-processing:} We report performance on three possible combinations of the data features as input:
\begin{enumerate}
    \item Vision + F/T + Force (V)
    \item Tactile + F/T + Force (T)
    \item Vision + Tactile + F/T + Force (V + T)
\end{enumerate}
We avoid using datapoints where the object was dropped during the grasp or pose phase, as in such cases the object would not be present during the shaking phase.  Unless otherwise specified, we use 20 timesteps spaced evenly from the start of the grasp to the end of the pose phase. We standardize each coordinate of the input features to have a mean of 0 and a standard deviation of 1 over all examples and timesteps.

\subsection{Improved classification using Deferred-Resampling} \label{drs}

It is possible that the labels during the pose phase and the shaking phase can differ, because an object can be stable when moving to the pose but unstable when shaking in the holding pose (or vice-versa). Examples, where these labels differ are particularly challenging because accurate classification requires predicting a different label than the one currently reflected in the sensor readings. This difficulty is also compounded by the fact only $\approx$20\% of examples have different labels during the pose and shaking phases. 

We propose to improve performance on such examples by adapting techniques used for learning with dataset imbalance which under-sample the majority class~\cite{japkowicz2002class,buda2018systematic,cao2019learning}. For our setting, the source of imbalance is not from the label distribution, but rather, whether an example has the same label in the pose and shaking phases. Overall, the Deferred-Resampling method~\cite{cao2019learning}~in this section improves test accuracy on unseen objects by 2.63\% over the vanilla LSTM baseline.

More details about the Deferred-Resampling method can be found in the appendix. 
%%%%%%%%%%%%%%%%%%%%%%%%%%%%%%%%%%%%%%%%%%%%%%%%%%%%%%%%%%%%%%%%%%%%%%%%%%%%%%%%

%%%%%%%%%%%%%%%%%%%%%%%%%%%%%%%%%%%%%%%%%%%%%%%%%%%%%%%%%%%%%%%%%%%%%%%%%%%%%%%%
\section{Experiments and Discussion}
Stability classification requires understanding the dynamics at different holding poses, thus making it a good probing task as the first step to modeling how humans grasp objects in the real world. We train models to use sensor readings from the grasp and pose phase to predict stability during the shaking phase and evaluate accuracy on held-out test sets. We find that training an LSTM with DRS on Tactile + Vision + F/T + Force value data performs best, achieving 85.21\% accuracy when train and test come from the same distribution, showing the usefulness of the multi-modal nature of~\poseit. (See Table~\ref{tab:unseenposes}.) 

The contributions of this section are:
\begin{enumerate}
    \item We show that although the classifier can generalize to new holding poses, test accuracy suffers when the test and train poses are very different. This demonstrates the value of data from diverse holding poses.
    \item We show that our models generalize to new poses and objects. This points to the potential for pre-training grasp stability models using~\poseit, which could be finetuned and deployed for real-world robotic grasping tasks.
\end{enumerate}

{\noindent \bf Classifiers:} The subsequent sections will frequently refer to the following classifiers and algorithms: 

{\it LSTM:} This is the vanilla architecture described in Section~\ref{model_arch}. 

{\it LSTM+DRS:} We train the LSTM described in Section~\ref{model_arch} using the Deferred-Resampling (DRS) technique (Section~\ref{drs}). DRS ensures that the model places more emphasis on learning from examples where the label changed between pose and shaking phases.

 All implementation details are deferred to the Appendix.

\subsection{Generalization to unseen poses} \label{sec:unseen_poses}

\begin{table}[t]
\centering
{\small
}
\begin{tabular}{@{}cccc@{}}
\toprule
Features                                  & Pose Group & \begin{tabular}[c]{@{}c@{}}Random \\ Poses (5 test)\end{tabular} & \begin{tabular}[c]{@{}c@{}}Uniform \\ Random Split\end{tabular} \\ \midrule
Tactile                                                        & 81.55\%    & 83.85\%                                                     & 84.52\%                                                         \\ \midrule
\begin{tabular}[c]{@{}c@{}}Vision + Tactile\end{tabular} & 82.4\%     & 84.28\%                                                     & \textbf{85.21\%}                                                \\ \bottomrule
\end{tabular}
\caption{\textbf{Generalization to unseen poses:} We display test accuracy results for 3 different methods of splitting holding poses into train and test. The model obtains 85.21\% accuracy when the same poses are in train and test. It performs worst when tested on poses than it was not trained on, as the train and test distributions are least similar in this setting. This demonstrates the value of collecting data for a large and diverse set of holding poses for each object.}
\label{tab:unseenposes}
\end{table}

In this section, we demonstrate the value of having data from diverse holding poses. We test the generalization of the classifier to unseen poses and verify that generalization performance is improved when similar poses are in the train and test sets. To conduct this experiment, we split the dataset into train and test in 3 different ways: 

{\it Uniform random split:} We randomly partition the data into train and test. Thus, the train and test sets are drawn from the same distribution, with no difference between object or pose. To keep the training set size consistent with the dataset splits below, we put 68.75\% of the data cycles in the training set. To reduce variance, we obtain 15 different splits of the dataset and report the average test accuracy of classifiers trained on these 15 splits of the dataset. 

{\it Random 5 poses:} We randomly select 5 of the 16 poses to put in the test set and partition data corresponding to all other poses into the training set. To reduce variance, we split the dataset this way 15 different times and report average test accuracy over classifiers trained on each split of the dataset. 

{\it Pose group:} Recall that poses 2-16 consist of 3 groups of 5 similar poses each. We put 1 group in the test set and the other 2 groups in the train set. We report average test results over all choices of the training set, where for each choice we train 5 classifiers (with different random seeds) to reduce variance. 

\noindent{\bf Results:} We report results for LSTM+DRS in Table~\ref{tab:unseenposes}. When we split train and test by the pose group, the classifier performs the worst on the test set, as the poses used in training and testing are the least similar. The classifier trained on the uniform random split of the data performs best, achieving 85.21\% accuracy. This demonstrates clear value in having a dataset with diverse held poses for each object. 

\subsection{Generalization to unseen objects}\label{sec:new_obj}
In this section, we analyze the ability of our classifiers trained on tactile data to generalize to unseen objects. We provide a baseline that uses a proxy label of whether the object slips during the \textit{pose} phase. This baseline performs poorly, showing that checking stability after the object moves to the holding pose is necessary for our task. 

As our dataset consists of data gathered from 26 objects, we use 22 objects in the training set and test generalization to the unseen test set of 4 objects. To reduce the variance over the choice of objects in the train and test set, we perform experiments on 20 randomly selected combinations of 4 objects for the test set. We report average numbers over the 20 ways of splitting train and test and 5 different random seeds for training the classifier. We compare our approaches against the following baseline/ceiling classifiers:

 {\it Majority classifier: }This trivial classifier labels all examples as either stable or not stable, depending on the majority. 

 {\it LSTM pose label (LSTM-P): }This baseline LSTM is trained on the stability label for the \textit{pose} phase. 

{\it LSTM whole cycle (LSTM-WC): }This classifier is trained on 20 evenly spaced sensor readings from the start of grasping to releasing. We expect this classifier to outperform others because it has access to data from the shaking phase, so it only needs to detect, rather than predict grasp failure.

\begin{table}[t]
\centering
\begin{tabular}{@{}lccc@{}}
\toprule \addlinespace[0.15cm]
Classifier          & Vision       & Tactile       & Both              \\ \midrule
\textit{LSTM-WC (Ceiling)}   & \textit{67.31} & \textit{77.89} & \textit{80.8}           \\
         Majority Classifier (Baseline) & 63.21 & 63.21 & 63.21          \\ 
         LSTM-P (Baseline)   & 64.47 & 66.57 & 66.43          \\
         Linear (Baseline)   & 63.31 & 64.22 & 66.57          \\
         LSTM (Ours)         & 66.2  & 73.45 & 74.66          \\
         LSTM+DRS (Ours)     & \textbf{68.25} & \textbf{74.76} & \textbf{77.29} \\ \bottomrule
\end{tabular}
{\small
\caption{\textbf{Generalization to new objects with tactile data:}  We show the test accuracy of our classifiers on three different combinations of modalities. Our described model to predict stability in unseen objects (LSTM+DRS) outperforms baselines. Using all of the modalities (Vision+Tactile+F/T+Force) produces the highest accuracy. For reference, we also include LSTM-WC, a ceiling that uses sensory data from the entire data collection cycle.}\label{tab:gelsightresults}} 
\end{table}

\noindent{\bf Results:}
We show our results in Table~\ref{tab:gelsightresults}. 
Our LSTM+DRS classifier produces the highest accuracy across all different variations of data input, besides the LSTM-WC whole cycle ceiling (which we expect to perform best, given it has access to data across all timesteps). Additionally, using all four modalities (RGB, Tactile, F/T, and Force) together outperforms using a subset of those features. 

We also note that the pose label baseline only performs 3\% better than the majority classifier, and performs much worse than the DRS classifier. This demonstrates the importance of the labeled data from the shaking phase, as simply knowing the label from the pose phase is \textit{not} sufficient. 
 
%%%%%%%%%%%%%%%%%%%%%%%%%%%%%%%%%%%%%%%%%%%%%%%%%%%%%%%%%%%%%%%%%%%%%%%%%%%%%%%%

%%%%%%%%%%%%%%%%%%%%%%%%%%%%%%%%%%%%%%%%%%%%%%%%%%%%%%%%%%%%%%%%%%%%%%%%%%%%%%%%
\section{Conclusion and Future Work}
We propose \textit{PoseIt}, a dataset that contains visual and tactile data from grasping objects and moving them to a holding pose for a stability check. This data is the next step in modeling how humans interact with objects in the real world: humans want to hold objects at various stable poses, rather than just lifting them vertically. Our experiments show that \textit{PoseIt} provides unique data that allows us to predict grasp stability at specific holding poses.

For future work, we would like to take even further advantage of \textit{PoseIt} to tackle the question of re-positioning an object to more stable holding poses. We would like to directly predict which poses are stable, using only sensor readings taken from a single reference pose.
%%%%%%%%%%%%%%%%%%%%%%%%%%%%%%%%%%%%%%%%%%%%%%%%%%%%%%%%%%%%%%%%%%%%%%%%%%%%%%%%

%%%%%%%%%%%%%%%%%%%%%%%%%%%%%%%%%%%%%%%%%%%%%%%%%%%%%%%%%%%%%%%%%%%%%%%%%%%%%%%%
\section*{APPENDIX}
{\bf Robot setup:}
See Figure~\ref{fig:data_modalities} for a visualization of the setup and data modalities. We use a Universal Robotics UR5e 6-DoF robot arm. An OnRobot Hex 6-Axis force/torque (F/T) sensor is attached to the end effector and records F/T measurements of the grasped objects. We use a high resolution visuo-tactile GelSight sensor. We use the Servo-electric 2-finger parallel gripper WSG50 from Weiss Robotics to attach these tactile sensors. We utilize two Azure Kinects and one RGB camera to visually capture the robot's workspace. 

{\bf Deferred-Resampling method:}
Let $S_=, S_{\ne}$ be the set of training examples where the pose and shaking phase labels match and differ, respectively. Let $r \triangleq \frac{|S_{\ne}|}{S_=}$ denote the ratio of these set sizes, i.e. in our case $r \approx 0.25$, and let $\sigma$ denote a desired sampling ratio, where $\sigma > r$ and $\sigma$ is a hyperparameter which we tune, from the choices $\{0.5, 1\}$. Our resampling scheme obtains each batch $B$ of examples for SGD updates as follows: 
\begin{enumerate}
    \item Sample a set $\tilde{B}$ of examples, for fixed $|\tilde{B}|$.
    \item For each $x \in \tilde{B} \cap S_=$, add $x$ to the set $B_=$ with probability $r/\sigma$.
    \item Output batch $B \triangleq B_= \cup (\tilde{B} \cap S_{\ne})$.
\end{enumerate}
This ensures that the ratio of SGD updates on examples from $S_{\ne}$ v.s. examples from $S_=$ equals $\sigma$ in expectation. 
Following the Deferred-Resampling (DRS) method of~\cite{cao2019learning}, we don't resample before annealing the SGD learning rate. This improves the quality of learned low-level features.

{\bf Implementation details:}
We split non-training data into a validation set and test set using a 50/50 random split and report test set numbers corresponding to the best validation performance. This mitigates fast over-fitting when the test data contains unseen objects.

\noindent{\bf Implementation details for Section~\ref{sec:unseen_poses}:}\label{sec:unseen_pose_app} We train our models using stochastic gradient descent (SGD) with initial learning rate of $0.01$, weight decay of $0.01$, dropout probability of $0.1$, hidden dimension of $500$. We train the models for 600 iterations with annealing at the 300-th iteration. For LSTM+DRS, we initially sample a batch of size 200 before further undersampling examples where the pose and stability check labels match. Our sampling probability is chosen according to the hyperparameter $\sigma$ defined in more detail in Section~\ref{drs}. We use $\sigma=0.5$ for the (T) model and $\sigma=1$ for the (V) and (V + T) models. 

\noindent{\bf Implementation details for Section~\ref{sec:new_obj}:}\label{sec:new_obj_app}
For the experiments in Section~\ref{sec:new_obj_app} section, we train for 500 iterations with annealing at the 30th iteration. For models trained without DRS, the batch size is 200. All other hyperparameters are the same as above. 
%%%%%%%%%%%%%%%%%%%%%%%%%%%%%%%%%%%%%%%%%%%%%%%%%%%%%%%%%%%%%%%%%%%%%%%%%%%%%%%%

%%%%%%%%%%%%%%%%%%%%%%%%%%%%%%%%%%%%%%%%%%%%%%%%%%%%%%%%%%%%%%%%%%%%%%%%%%%%%%%%

%%%%%%%%%%%%%%%%%%%%%%%%%%%%%%%%%%%%%%%%%%%%%%%%%%%%%%%%%%%%%%%%%%%%%%%%%%%%%%%%

\end{document}